\documentclass[letter, 10pt, conference]{ieeeconf}  
\usepackage[cmex10]{amsmath}
\usepackage{mathtools}
\usepackage{tabularx,booktabs}
\usepackage{graphicx}
\usepackage{array}
\usepackage[utf8]{inputenc}
\usepackage{tabu}
\usepackage{tabularx}
\usepackage{booktabs}
\usepackage[english]{babel}
\usepackage{csquotes}
\usepackage[skip=2pt,font=small]{caption}
\usepackage{footnote}
\usepackage{dblfloatfix}
\usepackage{amssymb}
\usepackage{verbatim}
\usepackage{multirow}
\usepackage{dsfont}
\usepackage{textcomp}
\usepackage{siunitx}
\usepackage{xcolor}
\usepackage{resizegather}
\usepackage{paralist}
\usepackage[mathscr]{euscript}

\newcommand {\zk}[1]{{\color{blue}\textbf{Zsolt: }#1}\normalfont}

\newcommand{\switchzk}[1]{%
  \ifthenelse{\equal{#1}{0}}{\renewcommand{\zk}[1]{}}{}}

\newcommand\newblock{\hskip .11em\@plus.33em\@minus.07em}
\def\@fnsymbol#1{\ensuremath{\ifcase#1\or *\or \dagger\or \ddagger\or
   \mathsection\or \mathparagraph\or \|\or **\or \dagger\dagger
   \or \ddagger\ddagger \else\@ctrerr\fi}}
\usepackage{url}
\usepackage[square,numbers,comma, sort]{natbib}
\RequirePackage{color}
\usepackage[colorlinks]{hyperref}
\ifdefined\nohyperref\else\ifdefined\hypersetup
  \definecolor{mydarkblue}{rgb}{0,0.08,0.45}
  \hypersetup{ %
    pdftitle={},
    pdfauthor={},
    pdfsubject={},
    pdfkeywords={},
    pdfborder=0 0 0,
    pdfpagemode=UseNone,
    colorlinks=true,
    linkcolor=mydarkblue,
    citecolor=mydarkblue,
    filecolor=mydarkblue,
    urlcolor=magenta,
    pdfview=FitH}
\DeclareQuoteStyle[american]{english}
    {\itshape\textquotedblleft}
    [\textquotedblleft]
    {\textquotedblright}
    [0.05em]
    {\textquoteleft}
    {\textquoteright}
\DeclareUnicodeCharacter{2217}{-}
\graphicspath{ {images/} }
\usepackage[linesnumbered, ruled]{algorithm2e}

\IEEEoverridecommandlockouts                              % This command is only needed if 
\renewcommand{\footnoterule}{%
  \hrule width 2in
  \kern 2pt
}

\definecolor{mycolor}{RGB}{147,112,219}

\begin{document}

\title{\LARGE \bf CenterSnap: Single-Shot Multi-Object 3D Shape Reconstruction \\ and
Categorical 6D Pose and Size Estimation}

\author{Muhammad Zubair Irshad\textsuperscript{*$\dag$}, Thomas Kollar\textsuperscript{$\ddag$}, Michael Laskey\textsuperscript{$\ddag$}, Kevin Stone\textsuperscript{$\ddag$}, Zsolt Kira\textsuperscript{$\dag$}
}
\maketitle
{
  \renewcommand{\thefootnote}%
    {\fnsymbol{footnote}}
\footnotetext[1]{Work done while author was research intern at Toyota Research Institute}    
  \footnotetext[2]{Georgia Institute of Technology \tt\scriptsize~(mirshad7, zkira)@gatech.edu}
    \footnotetext[3]{Toyota Research Institute \tt\scriptsize~(firstname.lastname)@tri.global}
}
\begin{comment}
Recently proposed navigation graph based discrete VLN settings become a poor proxy for real-world robotics navigation.
\end{comment}

\begin{abstract}
This paper studies the complex task of simultaneous multi-object 3D reconstruction, 6D pose and size estimation from a single-view RGB-D observation. In contrast to~\textit{instance-level} pose estimation, we focus on a more challenging problem where CAD models are not available at inference time. Existing approaches mainly follow a complex multi-stage pipeline which first localizes and detects each object instance in the image and then regresses to either their 3D meshes or 6D poses. These approaches suffer from high-computational cost and low performance in complex multi-object scenarios, where occlusions can be present. Hence, we present a simple one-stage approach to predict both the 3D shape and estimate the 6D pose and size jointly in a bounding-box free manner. In particular, our method treats object instances as spatial centers where each center denotes the complete shape of an object along with its 6D pose and size. Through this per-pixel representation, our approach can reconstruct in real-time~(40 FPS) multiple novel object instances and predict their 6D pose and sizes in a single-forward pass. Through extensive experiments, we demonstrate that our approach significantly outperforms all shape completion and categorical 6D pose and size estimation baselines on multi-object ShapeNet and NOCS datasets respectively with a 12.6$\percent$ absolute improvement in mAP for 6D pose for novel real-world object instances.
\end{abstract}
% no keywords
% For peer review papers, you can put extra information on the cover
% page as needed:
% \ifCLASSOPTIONpeerreview
% \begin{center} \bfseries EDICS Category: 3-BBND \end{center}
% \fi
%
% For peerreview papers, this IEEEtran command inserts a page break and
% creates the second title. It will be ignored for other modes.
%\IEEEpeerreviewmaketitle
\section{Introduction}

Multi-object 3D shape reconstruction and  6D pose~(i.e. 3D orientation and position) and size estimation from raw visual observations is crucial for robotics manipulation~\cite{cifuentes2016probabilistic, jiang2021synergies, laskey2021simnet}, navigation~\cite{qi2018frustum, chen20153d} and scene understanding~\cite{zhang2021holistic, Nie_2020_CVPR}. The ability to perform pose estimation in real-time leads to fast feedback control~\cite{kappler2018real} and the capability to reconstruct complete 3D shapes~\cite{ kuo2020mask2cad, niemeyer2020differentiable, mescheder2019occupancy} results in fine-grained understanding of local geometry, often helpful in robotic grasping~\cite{jiang2021synergies, ferrari1992planning}. Recent advances in deep learning have enabled great progress in~\textit{instance-level} 6D pose estimation~\cite{kehl2017ssd, rad2017bb8, xiang2018posecnn} where the exact 3D models of objects and their sizes are known a-priori. Unfortunately, these methods~\cite{tekin2018real, peng2019pvnet, wang2019densefusion} do not generalize well to realistic-settings on novel object instances with unknown 3D models in the same category, often referred to as~\textit{category-level settings}. Despite progress in category-level pose estimation, this problem remains challenging even when similar object instances are provided as priors during training, due to a high variance of objects within a category.

\begin{figure}[t!]
\centering
\includegraphics[width=0.95\columnwidth]{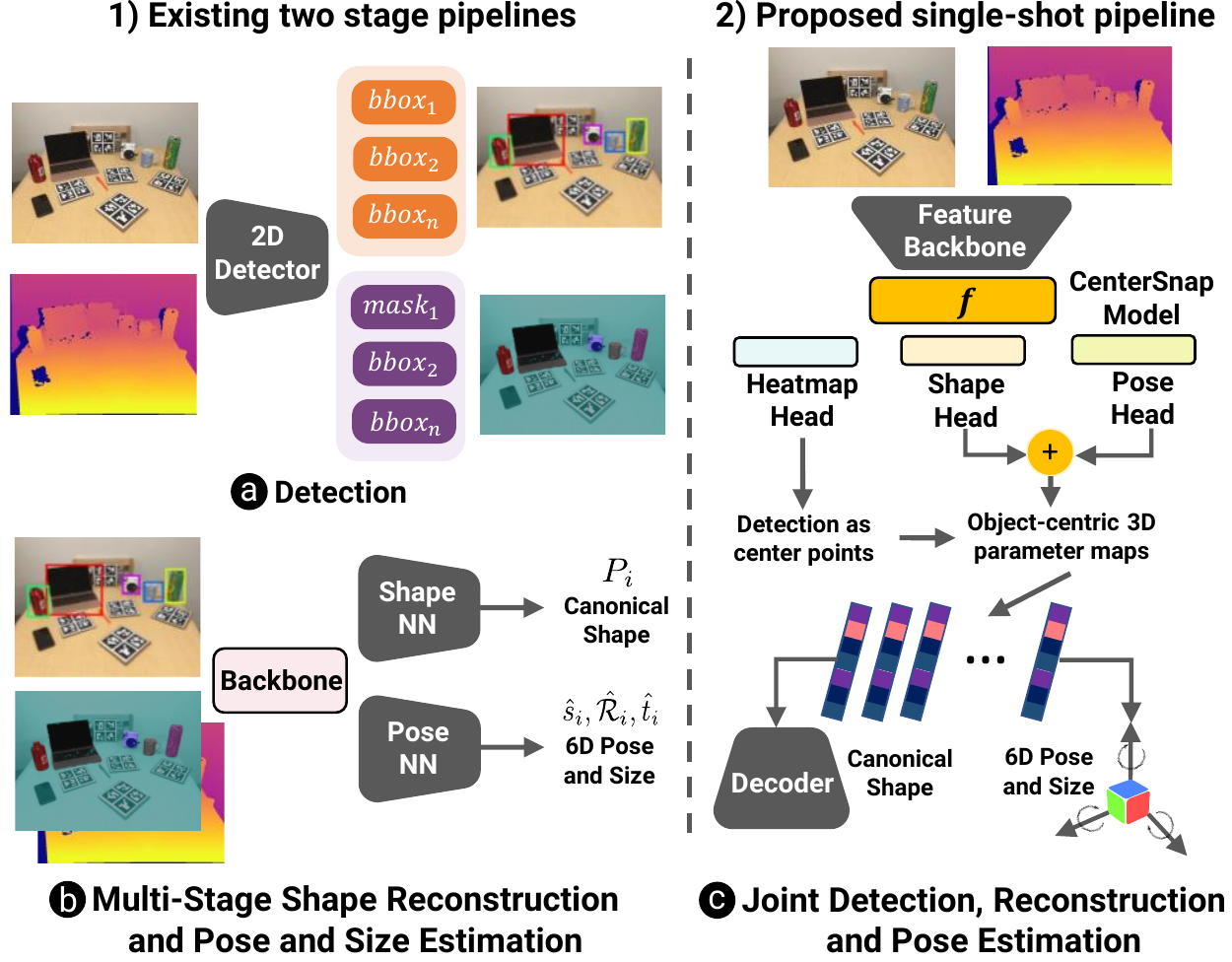}
\centering
  \caption{
  \textbf{Overview:}
    \textbf{(1)} Multi-stage pipelines in comparison to \textbf{(2)} our single-stage approach. The single-stage approach uses object instances as centers to jointly optimize 3D shape, 6D pose and size.
  }
  \label{ceterpoint_comparison}
\end{figure}

Recent works on shape reconstruction~\cite{gkioxari2019mesh, kato2018neural} and category-level 6D pose and size estimation~\cite{tian2020shape, wang2019normalized, sundermeyer2020augmented} use complex multi-stage pipelines. As shown in Figure~\ref{ceterpoint_comparison}, these approaches independently employ two stages, one for performing 2D detection~\cite{girshick14CVPR, he2017mask, ren2015faster} and another for performing object reconstruction or 6D pose and size estimation. This pipeline is computationally expensive, not scalable, and has low performance on real-world novel object instances, due to the inability to express explicit representation of shape variations within a category. Motivated by above, we propose to reconstruct complete 3D shapes and estimate 6D pose and sizes of novel object instances within a specific category, from a single-view RGB-D in a~\textit{single-shot manner}.\\
To address these challenges, we introduce Center-based Shape reconstruction and 6D pose and size estimation (CenterSnap), a single-shot approach to output complete 3D information (3D shape, 6D pose and sizes of multiple objects) in a bounding-box proposal-free and per-pixel manner. Our approach is inspired by recent success in anchor-free, single-shot 2D key-point estimation and object detection~\cite{nie2019single, zhou2019objects, zhou2020tracking, duan2019centernet}. As shown in Figure~\ref{ceterpoint_comparison}, we propose to learn a spatial per-pixel representation of multiple objects at their center locations using a feature pyramid backbone~\cite{laskey2021simnet, girshick14CVPR}. Our technique directly regresses multiple shape, pose, and size codes, which we denote as \textit{object-centric 3D parameter maps}. At each object's center point in these spatial object-centric 3D parameter maps, we predict vectors denoting the complete 3D information~(i.e. encoding 3D shape, 6D pose and sizes codes). 
A 3D auto-encoder~\cite{fan2017point, park2019deepsdf} is designed to learn canonical shape codes from a large database of shapes. A joint optimization for detection, reconstruction and 6D pose and sizes for each object's spatial center is then carried out using learnt shape priors. Hence, we perform complete 3D scene-reconstruction and predict 6D pose and sizes of novel object instances in a single-forward pass, foregoing the need for complex multi-stage pipelines~\cite{girshick14CVPR, gkioxari2019mesh, wang2019normalized}.

Our proposed method leverages a simpler and computationally efficient pipeline for a complete object-centric 3D understanding of multiple objects from a single-view RGB-D observation. We make the following contributions:
\begin{itemize}
\item Present the first work to formulate object-centric~\textit{holistic scene-understanding}~(i.e. 3D shape reconstruction and 6D pose and size estimation) for multiple objects from a single-view RGB-D in a~\textit{single-shot manner}. 
\item Propose a fast~(real-time) joint reconstruction and pose estimation system. Our network runs at 40 FPS on a NVIDIA Quadro RTX 5000 GPU.
\item Our method significantly outperforms all baselines for 6D pose and size estimation on NOCS benchmark, with over $12\percent$ absolute improvement in mAP for 6D pose.
\end{itemize}
\section{Related Work}~\label{sec:A}
\textbf{3D shape prediction and completion:} 3D reconstruction from a single-view observation has seen great progress with various input modalities studied. RGB-based shape reconstruction~\cite{fan2017point, choy20163d, groueix2018} has been studied to output either pointclouds, voxels or meshes~\cite{park2019deepsdf, mescheder2019occupancy, chen2019learning}. Contrarily, learning-based 3D shape completion~\cite{Yang18, varley2017shape, yuan2018pcn} studies the problem of completing partial pointclouds obtained from masked depth maps. However, all these works focus on reconstructing a single object. In contrast, our work focuses on multi-object reconstruction from a single RGB-D. Recently, multi-object reconstruction from RGB-D has been studied~\cite{engelmann2021points, runz2020frodo, gkioxari2019mesh}. However, these approaches employ complex multi-stage pipelines employing 2D detections  and then predicting canonical shapes. Our approach is a simple, bounding-box proposal-free method which jointly optimizes for detection, shape reconstruction and 6D pose and size. 

\textbf{Instance-Level and Category-Level 6D Pose and Size Estimation:} Works on Instance-level pose estimation use classical techniques such as template matching~\cite{kehl2016deep, sundermeyer2018implicit, tejani2014latent}, direct pose estimation~\cite{kehl2017ssd, wang2019densefusion, xiang2018posecnn} or point correspondences~\cite{tekin2018real, rad2017bb8}. Contrarily, our work closely follows the paradigm of category-level pose and size estimation where CAD models are not available during inference. Previous work has employed complex multi-stage pipelines~\cite{wang2019normalized,tian2020shape, chen2020learning} for category-level pose estimation. Our work optimizes for shape, pose, and sizes jointly, while leveraging the shape priors obtained by training a large dataset of CAD models. CenterSnap is a simpler, more effective, and faster solution.   
\begin{figure*}[t!]
\centering
\includegraphics[width=0.99\textwidth]{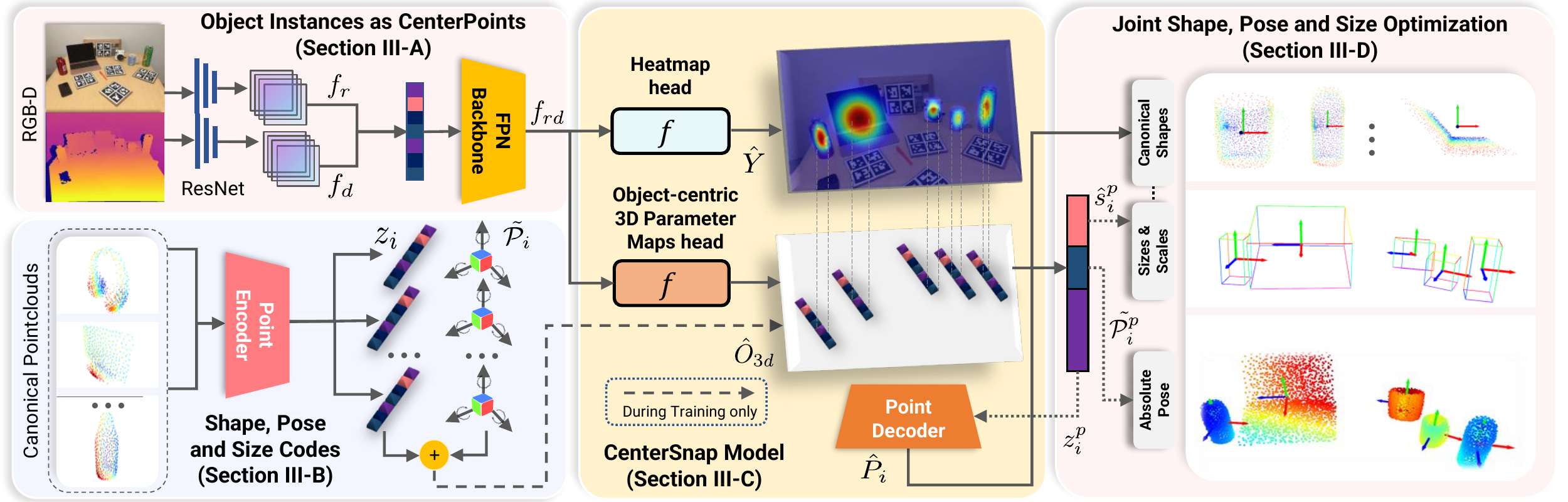}
\captionof{figure}{
\textbf{CenterSnap Method:} Given a single-view RGB-D observation, our proposed approach jointly optimizes for shape, pose, and sizes of each object in a single-shot manner. Our method comprises a joint FPN backbone for feature extraction~(Section \ref{backbone}), a pointcloud auto-encoder to extract shape codes from a large collection of CAD models~(Section~\ref{shapecode}), CenterSnap model which constitutes multiple specialized heads for heatmap and object-centric 3D parameter map prediction~(Section~\ref{centermodel}) and joint optimization for shape, pose, and sizes for each object's spatial center~(Section \ref{optimize}).
}
\vspace*{-7mm}
\label{framework}
\end{figure*}
\textbf{Per-pixel point-based representation} has been effective for anchor-free object detection and segmentation. These approaches~\cite{zhou2019objects, duan2019centernet, wang2020centermask} represent instances as their centers in a spatial 2D grid. This representation has been further studied for key-point detection~\cite{nie2019single}, segmentation~\cite{tian2020conditional,wang2020solo} and body-mesh recovery~\cite{sun2020centerhmr, ROMP}. Our approach falls in a similar paradigm and further adds a novelty to reconstruct object-centric holistic 3D information in an anchor-free manner. Different from~\cite{engelmann2021points, sun2020centerhmr}, our approach 1) considers pre-trained shape priors on a large collection of CAD models 2) jointly optimizes categorical shape 6D pose and size, instead of 3D-bounding boxes and 3) considers more complicated scenarios~(such as occlusions, a large variety of objects and sim2real transfer with limited real-world supervision).
\section{CenterSnap: Single-shot Object-Centric Scene Understanding of Multiple-objects}\label{centerpoint}

Given an RGB-D image as input, our goal is to simultaneously detect, reconstruct and localize all unknown object instances in the 3D space. In essence, we regard shape reconstruction and pose and size estimation as a point-based representation problem where each object's complete 3D information is represented by its center point in the 2D spatial image. Formally, given an RGB-D single-view observation~($I$ $\in$ $\mathds{R}^{h_{o}\times w_{o}\times 3 }$, $D$ $\in$ $\mathds{R}^{h_{o}\times w_{o}}$) of width $w_{o}$ and height $h_{o}$, our aim is to reconstruct the complete pointclouds~($P$ $\in$ $\mathds{R}^{K\times N\times 3 }$) coupled with 6D pose and scales~($\tilde{\mathbf{\mathcal P}}$ $\in$ $SE(3)$, $\hat{s}$ $\in$ $\mathbb{R}^3$) of all object instances in the 3D scene, where K is the number of arbitrary objects in the scene and N denotes the number of points in the pointcloud. The pose~($\tilde{\mathbf{\mathcal P}}$ $\in$ $SE(3)$) of each object is denoted by a 3D rotation $\hat{\mathcal{R}}$ $\in$ $SO(3)$ and a translation $\hat{t}$ $\in$ $\mathbb{R}^3$. The 6D pose $\tilde{\mathbf{\mathcal P}}$, 3D size (spatial extent obtained from canonical pointclouds~$P$) and 1D scales~$\hat{s}$ completely defines the unknown object instances in 3D space with respect to the camera coordinate frame.
To achieve the above goal, we employ an end-to-end trainable method as illustrated in Figure~\ref{framework}. First, objects instances are detected as heatmaps in a per-pixel manner~(Section~\ref{backbone}) using a CenterSnap detection backbone based on feature pyramid networks~\cite{laskey2021simnet, lin2017feature}. Second, a joint shape, pose, and size code denoted by object-centric 3D parameter maps is predicted for detected object centers using specialized heads~(Section~\ref{centermodel}). Our pre-training of shape codes is described in Section~\ref{shapecode}. Lastly, 2D heatmaps and our novel object-centric 3D parameter maps are jointly optimized to predict shapes, pose and sizes in a single-forward pass~(Section~\ref{optimize}).
\subsection{Object instances as center points}
\label{backbone}
We represent each object instance by its 2D location in the spatial RGB image following~\cite{zhou2019objects, duan2019centernet}. Given a RGB-D observation~($I$ $\in$ $\mathds{R}^{h_{o}\times w_{o}\times 3 }$, $D$ $\in$ $\mathds{R}^{h_{o}\times w_{o}}$), we generate a low-resolution spatial feature representations $f_{r}$ $\in$ $\mathds{R}^{h_{o}/4\times w_{o}/4\times C_{s}}$ and $f_{d}$ $\in$ $\mathds{R}^{h_{o}/4\times w_{o}/4\times C_{s}}$ by using Resnet~\cite{he2016deep} stems, where $C_{s}=32$. We concatenate computed features $f_{r}$ and $f_{d}$ along the channel dimension before feeding it to Resnet18-FPN backbone~\cite{kirillov2019panoptic} to compute a~\textit{pyramid of features} ($f_{rd}$) with scales ranging from 1/8 to 1/2 resolution, where each pyramid
level has the same channel dimension~(i.e. 64). We use these combined features with a specialized heatmap head to predict object-based heatmaps~$\hat{Y} \in [0,1]^{\frac{h_{o}}{R} \times \frac{w_{o}}{R} \times 1}$ where $R$\,=\,$8$ denotes the heat-map down-sampling factor. Our specialized heatmap head design merges the semantic information from all FPN levels into one output~($\hat{Y}$). We use three upsampling stages followed by element-wise sum and $softmax$ to achieve this. This design allows our network to 1) capture multi-scale information and 2) encode features at higher resolution for effective reasoning at the per-pixel level. We train the network to predict ground-truth heatmaps~($Y$) by minimizing MSE loss, $\mathcal{L}_{\text{inst}}=\sum_{xyg}\left(\hat{Y}-Y\right)^{2}$. The Gaussian kernel~${Y_{xyg} = \exp\left(-\frac{(x-c_x)^2+(y- c_y)^2}{2\sigma^2}\right)}$ of each center in the ground truth heat-maps~($Y$) is relative to the scale-based standard deviation~$\sigma$ of each object, following~\cite{laskey2021simnet, zhou2019objects, duan2019centernet, law2018cornernet}. 

\subsection{Shape, Pose, and Size Codes}
\label{shapecode}
\begin{figure}[t!]
\centering
\includegraphics[width=0.97\columnwidth]{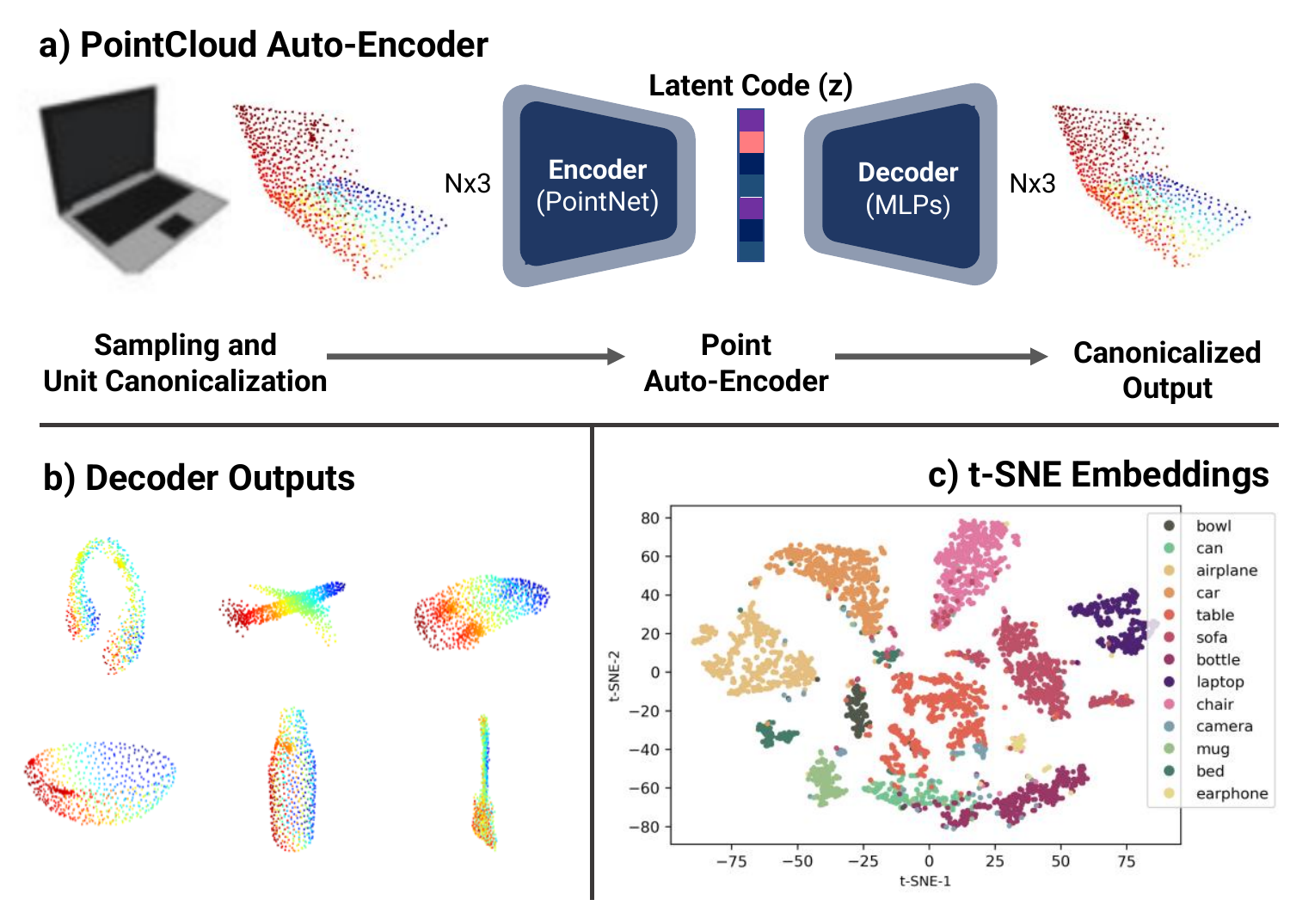}
\centering
  \caption{
  \textbf{Shape Auto-Encoder:} We design a Point Auto-encoder~(\textbf{a}) to find unique shape-code~($z_{i}$) for all the shapes. Unit-canonicalized pointcloud outputs from the decoder network are shown in~(\textbf{b}). t-SNE embeddings for shape-code~($z_{i}$) are visualized in~(\textbf{c})
  }
  \label{auto-encoder}
\end{figure}

\begin{figure}[t!]
\centering
\includegraphics[width=0.98\columnwidth]{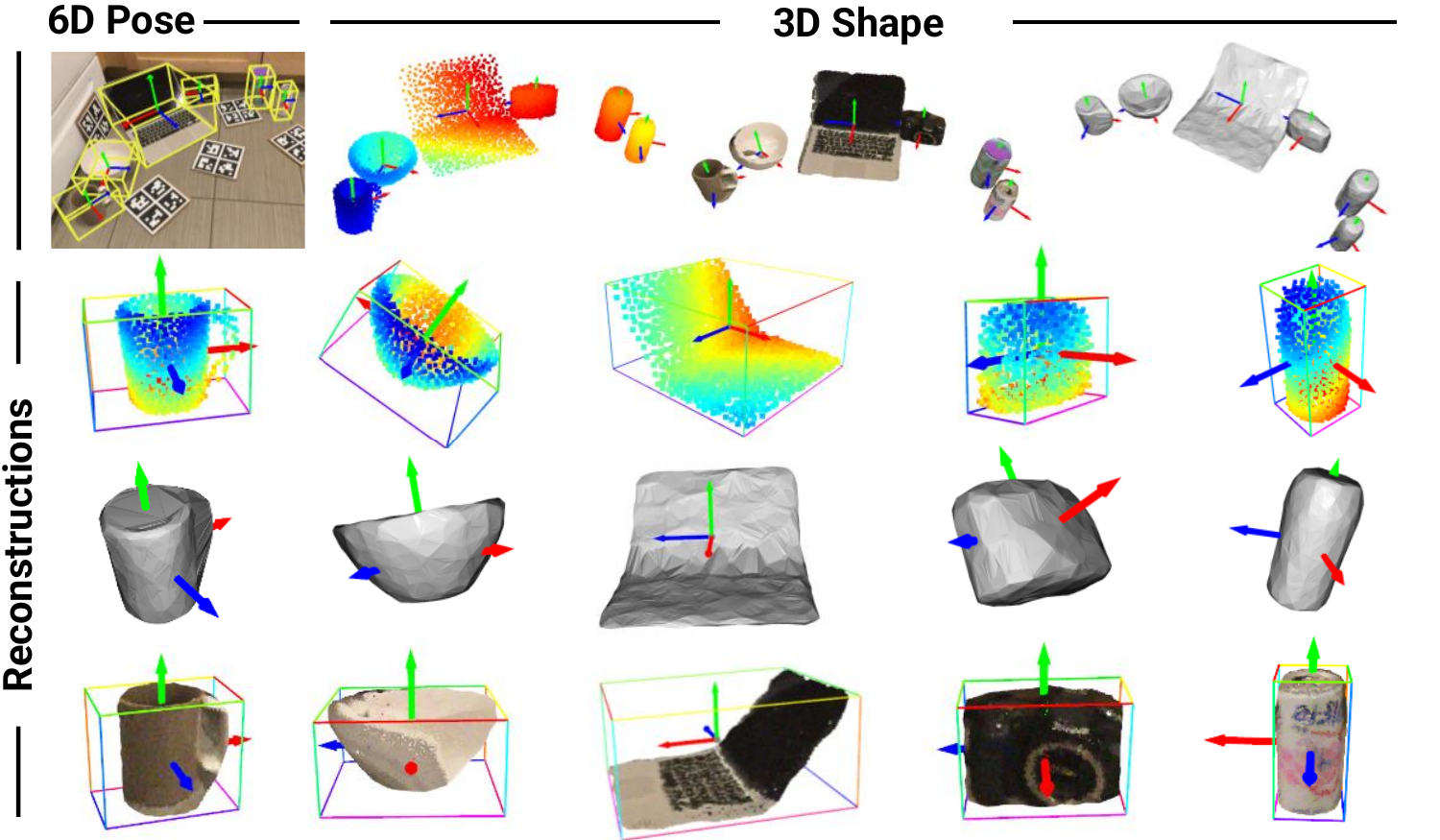}
\centering
  \caption{
  \textbf{Sim2Real Reconstruction:} Single-shot sim2real shape reconstructions on NOCS showing pointclouds, meshes and textures.
  }
  \label{reconstruction_qualitative}
\end{figure}

To jointly optimize the object-based heatmaps, 3D shapes and 6D pose and sizes, the complete object-based 3D information (i.e. Pointclouds $P$, 6D pose $\tilde{\mathbf{\mathcal P}}$ and scale $\hat{s}$) are represented as as object-centric 3D parameter maps~($O_{3d}$ $\in$ $\mathds{R}^{h_{o}\times w_{o}\times 141}$). $O_{3d}$ constitutes two parts, shape latent-code and 6D Pose and scales. The pointcloud representation for each object is stored in the object-centric 3D parameter maps as a latent-shape code~($z_{i}$ $\in$ $\mathbb{R}^{128}$). The ground-truth Pose~($\tilde{\mathbf{\mathcal P}}$) represented by a $3\times 3$ rotation $\hat{\mathcal{R}}$ $\in$ $SO(3)$ and translation $\hat{t}$ $\in$ $\mathbb{R}^3$ coupled with 1D scale $\hat{s}$ are vectorized to store in the $O_{3d}$ as 13-D vectors. To learn a shape-code~($z_{i}$) for each object, we design an auto-encoder trained on all 3D shapes from a set of CAD models. Our auto-encoder is representation-invariant and can work with any shape representation. Specifically, we design an encoder-decoder network (Figure~\ref{auto-encoder}), where we utilize a Point-Net encoder~($g_{\phi}$) similar to~\cite{qi2017pointnet}. The decoder network~($d_{\theta}$), which comprises three fully-connected layers, takes the encoded low-dimensional feature vector i.e. the latent shape-code ($z_{i}$) and reconstructs the input pointcloud $\hat{P}_{i} = d_{\theta}(g_{\phi}(P_{i}))$. To train the auto-encoder, we sample 2048 points from the ShapeNet~\cite{chang2015ShapeNet} CAD model repository and use them as ground-truth shapes. Furthermore, we unit-canonicalize the input pointclouds by applying a scaling transform to each shape such that the shape is centered at origin and unit normalized. We optimize the encoder and decoder networks jointly using the reconstruction-error, denoted by Chamfer-distance, as shown below:
\begin{equation} \label{eq:distance_Chamfer}
     D_{cd}(\mathbf{P}_{i}, \hat{\mathbf{P}}_{i}) = 
     \frac{1}{|\mathbf{P}_{i}|} \sum_{x \in \mathbf{P}_{i}} \min_{y \in \hat{\mathbf{P}}_{i}}\|x-y\|_{2}^{2} + \frac{1}{|\hat{\mathbf{P}}_{i}|}\sum_{\mathbf{y} \in \hat{\mathbf{P}}_{i}} \min_{x \in \mathbf{P}_{i}}\|x-y\|_{2}^{2} \nonumber
\end{equation}
Sample decoder outputs and t-SNE embeddings~\cite{van2008visualizing} for the latent shape-code~($z_{i}$) are shown in Figure~\ref{auto-encoder} and our complete 3D reconstructions on novel real-world object instances are visualizes in Figure~\ref{reconstruction_qualitative} as pointclouds, meshes and textures. Our shape-code space provides a compact way to encode 3D shape information from a large number of CAD models. As shown by the t-SNE embeddings (Figure~\ref{auto-encoder}), our shape-code space finds a distinctive 3D space for semantically similar objects and provides an effective way to scale shape prediction to a large number (i.e. 50+) of categories.
\subsection{CenterSnap Model}
\label{centermodel}
Given object center heatmaps~(Section~\ref{backbone}), the goal of the CenterSnap model is to infer object-centric 3D parameter maps which define each object instance completely in the 3D-space. The CenterSnap model comprises a task-specific head similar to the heatmap head~(Section~\ref{backbone}) with the input being \textit{pyramid of features}~($f_{rd}$). During training, the task-specific head outputs a 3D parameter map $\hat{O}_{3d}$ $\in$ $\mathds{R}^{\frac{h_{o}}{R}\times \frac{w_{o}}{R}\times 141}$ where each pixel in the down-sampled map~($\frac{h_{o}}{R}\times \frac{w_{o}}{R}$) contains the complete object-centric 3D information (i.e. shape-code $z_{i}$, 6D pose $\tilde{\mathbf{\mathcal P}}$ and scale $\hat{s}$) as 141-D vectors, where $R = 8$. Note that, during training, we obtain the ground-truth shape-codes from the pre-trained point-encoder $\hat{z}_{i} = g_{\phi}(P_{i})$. For Pose~($\tilde{\mathbf{\mathcal P}}$), our choice of rotation representation $\hat{\mathcal{R}}$ $\in$ $SO(3)$ is determined by stability during training~\cite{zhou2019continuity}. Furthermore, we project the predicted~$3\times3$ rotation $\hat{\mathcal{R}}$ into $SO(3)$, as follows: $\operatorname{SVD}^{+}(\hat{\mathcal{R}})=U \Sigma^{\prime} V^{T}, \text { where } \Sigma^{\prime}=\operatorname{diag}\left(1, 1, \operatorname{det}\left(U V^{T}\right)\right)$
To handle ambiguities caused by rotational symmetries, we also employ a rotation mapping function defined by~\cite{pitteri2019object}. The mapping function, used only for symmetric objects~\textit{(bottle, bowl, and can)}, maps ambiguous ground-truth rotations to a single canonical rotation by normalizing the pose rotation.
During training, we jointly optimize the predicted object-centric 3D parameter map~($\hat{O}_{3d}$) using a masked Huber loss~(Eq. \ref{huberloss}), where the Huber loss is enforced only where the Gaussian heatmaps ($Y$) have score greater than 0.3 to prevent ambiguity in areas where no objects exist. Similar to the Gaussian distribution of heatmaps in Section \ref{backbone}, the ground-truth \textit{Object 3D-maps}~($O_{3d}$) are calculated using the scale-based Gaussian kernel~$Y_{xyg}$ of each object. 
\begin{equation}
\small
\label{huberloss}
\resizebox{0.91\hsize}{!}{%
$\mathcal{L}_{3D}(O_{3d},\hat{O}_{3d})=\left\{\begin{array}{cc}
\frac{1}{2}(O_{3d}-\hat{O}_{3d})^{2} & if|(O_{3d}-\hat{O}_{3d})|<\delta \\
\delta\left((O_{3d}-\hat{O}_{3d})-\frac{1}{2} \delta\right) & \text {otherwise}
\end{array}\right.$}
\end{equation}

\textbf{Auxiliary Depth-Loss}: We additionally integrate an auxiliary depth reconstruction loss $\mathcal{L}_{D}$ for effective sim2real transfer, where $\mathcal{L}_{D}(D, \hat{D})$ minimizes the Huber loss (Eq. \ref{huberloss}) between target depth~($D$) and the predicted depth~($\hat{D}$) from the output of task-specific head, similar to the one used in Section~\ref{backbone}. The depth auxiliary loss (further investigated empirically using ablation study in Section~\ref{experiments&results}) forces the network to learn geometric features by reconstructing artifact-free depth. Since real depth sensors contain artifacts, we enforce this loss by pre-processing the input synthetic depth images to contain noise and random eclipse dropouts~\cite{laskey2021simnet}.

\begin{table*}[t]
    \scriptsize
    \centering
    \renewcommand{\arraystretch}{1.3}
    \caption{
    \textbf{Quantitative comparison of 3D object detection and 6D pose estimation on NOCS}~\cite{wang2019normalized}: Comparison with strong baselines. Best results are highlighted in \textbf{bold}. $*$ denotes the method does not evaluate size and scale hence does not report IOU metric. For a fair comparison with other approaches, we report the per-class metrics using nocs-level class predictions. Note that the comparison results are either fair re-evaluations from the author's provided best checkpoints or reported from the original paper.
    % \zk{Change all instances of Center-SnP to CenterSnap to make it consistent}
    % Note that these baselines are reimplementations from with small changes (see Section~\ref{baselines} for further details\zk{How do we know that these were implemented properly and it's a fair comparison? Did you make sure it reproduced original results? Since you use run-time does the run-time match with published run-times they mention? }). 
    }
    \label{comparison_table}
    \vspace{+0.15cm}
    \resizebox{1.0\textwidth}{!}{
    \begin{tabular}{clcccccccccccc}
        \toprule
        & & \multicolumn{6}{c}{\textbf{CAMERA25}} & \multicolumn{6}{c}{\textbf{REAL275}} \\ 
        \cmidrule(r{0.1in}){3-8} \cmidrule(r{0.1in}){9-14}
        & {Method} & \textbf{IOU25} & \textbf{IOU50} & \textbf{5\textdegree \SI{5}{\cm}} & \textbf{5\textdegree \SI{10}{\cm}} & \textbf{10\textdegree \SI{5}{\cm}} & \textbf{10\textdegree \SI{10}{\cm}} & \textbf{IOU25} & \textbf{IOU50} & \textbf{5\textdegree \SI{5}{\cm}} & \textbf{5\textdegree \SI{10}{\cm}}& \textbf{10\textdegree \SI{5}{\cm}} & \textbf{10\textdegree \SI{10}{\cm}}\\
        \cmidrule(r{0.1in}){2-2}
        \cmidrule(r{0.1in}){3-8} \cmidrule(r{0.1in}){9-14}
        1 & {NOCS~\cite{wang2019normalized}}      &  91.1& 83.9 & 40.9  & 38.6 & 64.6 & 65.1 & \textbf{84.8} & 78.0 & 10.0 & 9.8 & 25.2 & 25.8\\
        2 & {Synthesis$^{*}$~\cite{chen2020category}} &  - & - & -  &- & - & - & - & - &  0.9 & 1.4 & 2.4 & 5.5 \\

        3 & {Metric Scale~\cite{lee2021category}}      &  \textbf{93.8}& 90.7 & 20.2  & 28.2 & 55.4 & 58.9 & 81.6 & 68.1 & 5.3 & 5.5 & 24.7 & 26.5 \\
        4 & {ShapePrior~\cite{tian2020shape}} &81.6	&72.4&	59.0&	59.6&  81.0 &  81.3 &	81.2& 77.3	&	21.4	&21.4&	54.1&	54.1\\
        5 & {CASS~\cite{chen2020learning}} & - & - & - & - & - & - & 84.2 & 77.7 &  23.5 & 23.8 & 58.0 & 58.3\\
        \midrule
        6 & {\textbf{CenterSnap~(Ours)}} & 93.2&	92.3&	63.0	& 69.5 &	79.5 & 87.9&	83.5 &	80.2 & 27.2 &	29.2 & 58.8 &	64.4 \\
    7 & {\textbf{CenterSnap-R~(Ours)}} & 93.2&	\textbf{92.5}&	\textbf{66.2}	& \textbf{71.7} &	\textbf{81.3} & \textbf{87.9}&	83.5 &	\textbf{80.2} & \textbf{29.1} &	\textbf{31.6} & \textbf{64.3} &	\textbf{70.9} \\
        \bottomrule
    \end{tabular}
    }
    % \vspace{-2mm}
\end{table*}

\begin{table*}[t]
    \centering
    \scriptsize
    \renewcommand{\arraystretch}{1.3}
    \vspace{-1mm}
    \caption{
    \textbf{Quantitative comparison of 3D shape reconstruction on NOCS}~\cite{wang2019normalized}: Evaluated with \textbf{CD} metric ($10^{-2}).$ Lower is better.
    % Note that these baselines are reimplementations from VLN-CE~\cite{krantz2020navgraph} with small changes (see Section~\ref{baselines} for further details).\zk{Copied from prev. paper, needs update} 
    }
    
    \label{reconstruction_nocs}
    \resizebox{1.0\textwidth}{!}{
    \begin{tabular}{clcccccccccccccc}
        \toprule
        & & \multicolumn{7}{c}{\textbf{CAMERA25}} & \multicolumn{7}{c}{\textbf{REAL275}} \\ \cmidrule(r{0.1in}){3-9} \cmidrule(r{0.1in}){10-16}
        & {Method} & \textbf{Bottle} & \textbf{Bowl} & \textbf{Camera} & \textbf{Can} & \textbf{Laptop} & \textbf{Mug} & \textbf{Mean} & \textbf{Bottle} & \textbf{Bowl} & \textbf{Camera} & \textbf{Can} & \textbf{Laptop} & \textbf{Mug} & \textbf{Mean} \\
        \midrule
        1 & {Reconstruction~\cite{tian2020shape}} & 0.18 & 0.16  & 0.40 & 0.097 & 0.20 & 0.14 & 0.20& 0.34 &0.12 & 0.89 & 0.15 &0.29 &0.10 & 0.32 \\
        2 & {ShapePrior~\cite{tian2020shape}} &0.34 &0.22& 0.90& 0.22& 0.33& 0.21 &0.37 &0.50 &0.12& 0.99& 0.24& 0.71& 0.097& 0.44 \\
        \midrule
        3 & \textbf{CenterSnap~(Ours)} &0.11	&0.10&	0.29&	0.13&	0.07& 0.12	&	0.14 & 0.13 &  0.10 & 0.43 &0.09 & 0.07 & 0.06 & 0.15 \\
        \bottomrule
    \end{tabular}
    }
    \vspace{-5mm}
\end{table*}
\subsection{Joint Shape, Pose, and Size Optimization}
\label{optimize}
We jointly optimize for detection, reconstruction and localization. Specifically, we minimize a combination of heatmap instance detection, object-centric 3D map prediction and auxiliary depth losses as $\mathcal{L}= \lambda_{l}\mathcal{L}_{inst} + \lambda_{O_{3d}}\mathcal{L}_{O_{3d}} +  \lambda_{d}\mathcal{L}_{D}
$
where $\lambda$ is a weighting co-efficient with values determined empirically as 100, 1.0 and 1.0 respectively.

\textbf{Inference}:
During inference, we perform peak detection as in~\cite{zhou2019objects} on the heatmap output ($\hat{Y}$) to get detected centerpoints for each object, {$c_{i}$ $in$ $\mathbb{R}^2$} = ($x_{i}, {y_{i}}$) (as shown in Figure~\ref{framework}~\emph{middle}). These centerpoints are local maximum in heatmap output~($\hat{Y}$). We perform non-maximum suppression on the detected heatmap maximas using a~$3\times 3$ max-pooling, following~\cite{zhou2019objects}. Lastly, we directly sample the object-centric 3D parameter map of each object from~$\hat{O}_{3d}$ at the predicted center location~($c_{i}$) via~$\hat{O}_{3d}(x_{i}, {y_{i}})$. We perform inference on the extracted latent-codes using point-decoder to reconstruct pointclouds~($\hat{P}_{i} = d_{\theta}(z_{i}^{p})$). Finally, we extract~$3\times 3$ rotation $\hat{\mathcal{R}}_{i}^{p}$, 3D translation vector $\hat{t}_{i}^{p}$ and 1D scales $\hat{s}_{i}^{p}$ from~$\hat{O}_{3d}$ to get transformed points in the 3D space $\hat{P}_{i}^{recon} = [\hat{\mathcal{R}}_{i}^{p} | \hat{t}_{i}^{p}]*\hat{s}_{i}^{p}*\hat{P}_{i}$ (as shown in Figure~\ref{framework}~\emph{right-bottom}).
\section{Experiments \& Results} 
\label{experiments&results}
In this section, we aim to answer the following questions: 1) How well does CenterSnap reconstruct multiple objects from a single-view RGB-D observation? 2) Does CenterSnap perform fast pose-estimation in real-time for real-world applications? 3) How well does CenterSnap perform in terms of 6D pose and size estimation? 

\textbf{Datasets:} We utilize the \textbf{NOCS}~\cite{wang2019normalized} dataset to evaluate both shape reconstruction and categorical 6D pose and size estimation. We use the CAMERA dataset for training which contains 300K synthetic images, where 25K are held out for evaluation. Our training set comprises 1085 object instances from 6 different categories - \textit{bottle, bowl, camera, can, laptop and mug} whereas the evaluation set contains 184 different instances. The REAL dataset contains
4300 images from 7 different scenes for training, and 2750 real-world images from 6 scenes for evaluation. Further, we evaluate multi-object reconstruction and completion using Multi-Object ShapeNet Dataset \textbf{(MOS)}. We generate this dataset using the SimNet~\cite{laskey2021simnet} pipeline. Our datasets contains $640px \times 480px$ renderings of multiple~(3-10) ShapeNet objects~\cite{chang2015ShapeNet} in a table-top scene. Following~\cite{laskey2021simnet}, we randomize over lighting and textures using OpenGL shaders with PyRender~\cite{pyrender}.
Following~\cite{yuan2018pcn}, we utilize 30974 models from 8 different categories for training~(i.e. MOS-train): \textit{airplane, cabinet, car, chair, lamp, sofa, table}. We use the held out set (MOS-test) of 150 models for testing from a novel set of categories - \textit{bed, bench, bookshelf and bus}. \\
\textbf{Evaluation Metrics}: Following~\cite{wang2019normalized}, we independently evaluate the performance of 3D object detection and 6D pose estimation using the following key metrics: 1) Average-precision for various IOU-overlap thresholds (\textbf{IOU25} and \textbf{IOU50}). 2) Average precision of object instances for which the error is less than $n^{\circ}$ for rotation and $m$ cm for translation (\textbf{5\textdegree \SI{5}{\cm}}, \textbf{\textbf{5\textdegree \SI{10}{\cm}}} and \textbf{\textbf{10\textdegree \SI{10}{\cm}}}). For shape reconstruction we use Chamfer distance~(CD) following~\cite{yuan2018pcn}.\\
\textbf{Implementation Details:} CenterSnap is trained on the CAMERA training dataset with fine-tuning on the REAL training set. We use a batch-size of 32 and trained the network for 40 epochs with early-stopping based on the performance of the model on the held out validation set. We found data-augmentation~(i.e. color-jitter) on the real-training set to be helpful for stability and training performance. The auto-encoder network is comprised of a Point-Net encoder~\cite{qi2017pointnet} and three-layered fully-connected decoder each with output dimension of $512, 1024$ and $1024\times3$. The auto-encoder is frozen after initially training on CAMERA CAD models for 50 epochs. We use Pytorch~\cite{NEURIPS2019_9015} for all our models and training pipeline implementation. For shape-completion experiments, we train only on MOS-train with testing on MOS-test.\\
\begin{figure}[!b]
\centering
\includegraphics[width=0.90\columnwidth]{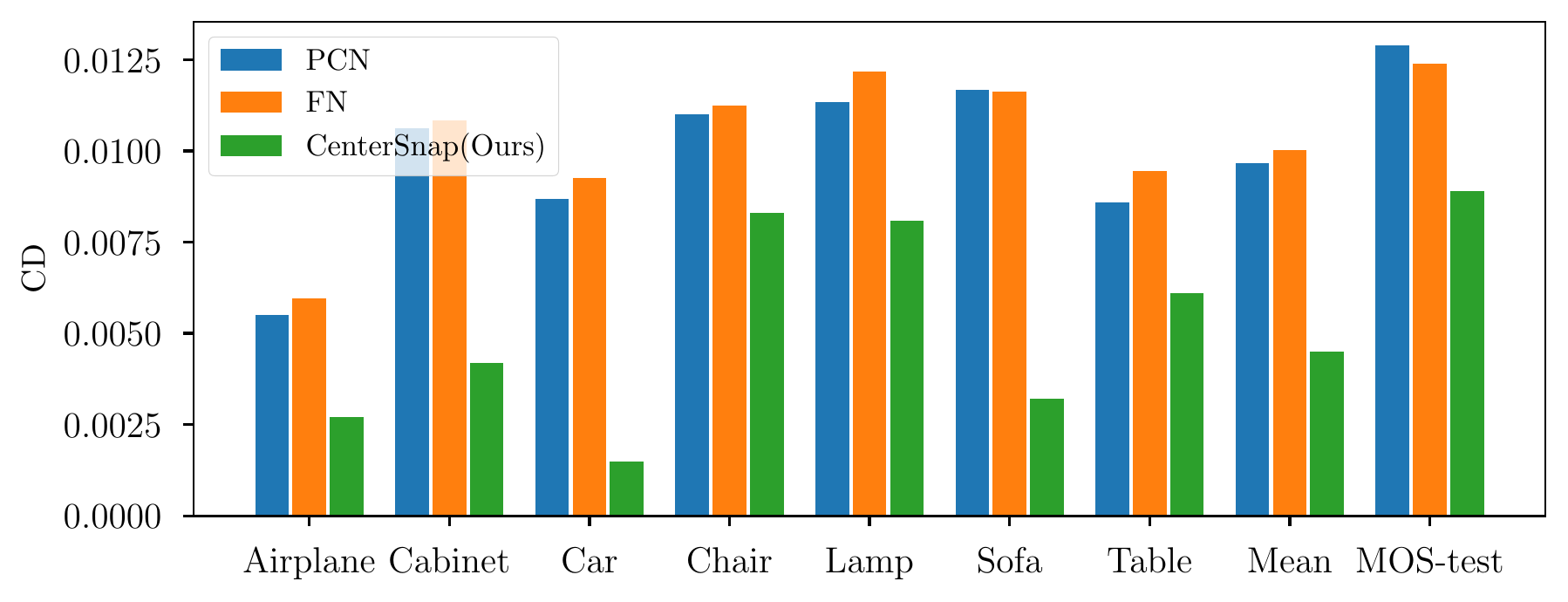}
\centering
  \caption{
  \textbf{Shape Completion:} Chamfer distance (CD reported on y-axis) evaluation on Multi-object ShapeNet dataset.
  }
  \label{shape_completion}
\end{figure}
\textbf{NOCS Baselines:}
We compare seven model variants to show effectiveness of our method:
(1) \textbf{NOCS}~\cite{wang2019normalized}: Extends Mask-RCNN architecture to predict NOCS map and uses similarity transform with depth to predict pose and size. Our results are compared against the best pose-estimation configuration in NOCS~(i.e. 32-bin classification) (2) \textbf{Shape Prior}~\cite{tian2020shape}: Infers 2D bounding-box for each object and predicts a shape-deformation. (3) \textbf{CASS}~\cite{chen2020learning}: Employs a 2-stage approach to first detect 2D bounding-boxes and second regress the pose and size. (4) \textbf{Metric-Scale}~\cite{lee2021category}: Extends NOCS to predict object center and metric shape separately (5) \textbf{CenterSnap:} Our single-shot approach with direct pose and shape regression. (6) \textbf{CenterSnap-R:} Our model with a standard point-to-plane iterative pose refinement~\cite{segal2009generalized, Zhou2018} between the projected canonical pointclouds in the 3D space and the depth-map.  Note that we do not include comparisons to 6D pose tracking baselines such as~\cite{wang20206, wen2021bundletrack} which are not detection-based~(i.e. do not report mAP metrics) and require pose initialization. \\
\begin{figure}[!b]
\centering
\includegraphics[width=0.9\columnwidth]{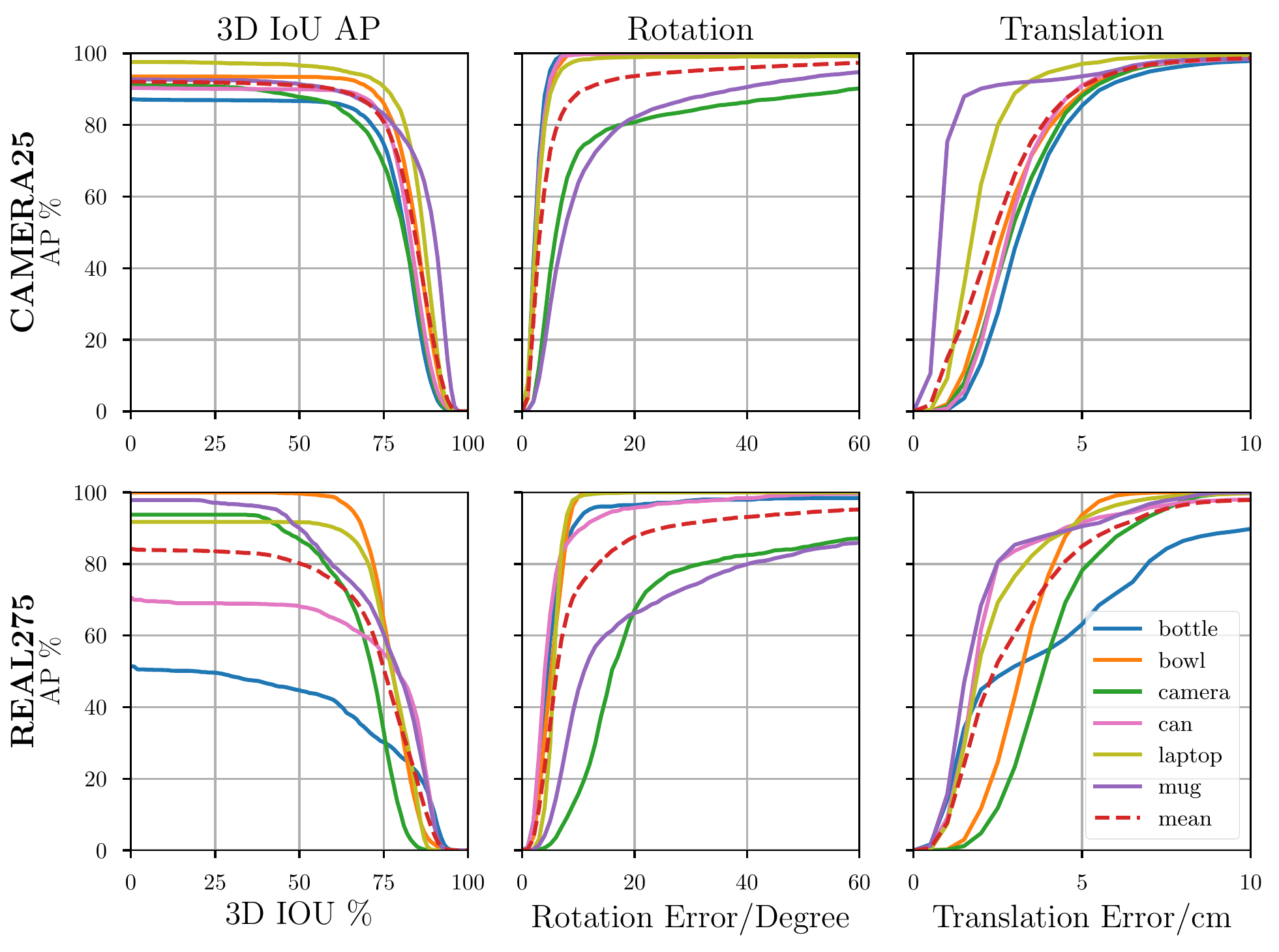}
\centering
  \caption{
  \textbf{mAP on Real-275 test-set:} Our method's mean Average Precision on NOCS for various IOU and pose error thresholds.
  }
  \label{nocs_resut}
\end{figure}
\textbf{Comparison with NOCS baselines:} The results of our proposed CenterSnap method are reported in Table~\ref{comparison_table} and Figure~\ref{nocs_resut}. Our proposed approach consistently outperforms all the baseline methods on both 3D object detection and 6D pose estimation. Among our variants, CenterSnap-R achieves the best performance. Our method~(i.e. CenterSnap) is able to outperform strong baselines (\#1 - \#5 in Table~\ref{comparison_table}) even without iterative refinement. Specifically, CenterSnap-R method shows superior performance on the REAL test-set by achieving a mAP of 80.2$\percent$ for 3D IOU at 0.5, 31.6$\percent$ for 6D pose at 5\textdegree \SI{10}{\cm} and 70.9$\percent$ for 6D pose at 10\textdegree \SI{10}{\cm}, hence demonstrating an absolute improvement of 2.7$\percent$, 10.8$\percent$ and 12.6$\percent$ over the best-performing baseline on the Real dataset.  Our method also achieves superior test-time performance on CAMERA evaluation never seen during training. We achieve a mAP of 92.5$\percent$ for 3D IOU at 0.5, 71.7$\percent$ for 6D pose at 5\textdegree \SI{10}{\cm} and 87.9$\percent$ for 6D pose at 10\textdegree \SI{10}{\cm}, demonstrating an absolute improvement of 1.8$\percent$, 12.1$\percent$ and 6.6$\percent$ over the best-performing baseline. \\
\textbf{NOCS Reconstruction:} To quantitatively analyze the reconstruction accuracy, we measure the Chamfer distance~(CD) of our reconstructed pointclouds with ground-truth CAD model in NOCS. Our results are reported in Table~\ref{reconstruction_nocs}. Our results show consistently lower CD metrics for all class categories which shows superior reconstruction performance on novel object instances. We report a lower mean Chamfer distance of 0.14 on CAMERA25 and 0.15 on REAL275 compared to 0.20 and 0.32 reported by the competitive baseline~\cite{tian2020shape}. \\ 
\textbf{Comparison with Shape Completion Baselines}: We further test our network’s ability to reconstruct complete 3D
shapes by comparing against depth-based shape-completion
baselines i.e. PCN~\cite{yuan2018pcn} and Folding-Net~\cite{yang2018foldingnet}. The results of
our CenterSnap method are reported in Figure 5. Our consistently lower Chamfer distance (CD) compared to strong
shape-completion baselines show our network’s ability to
reconstruct complete 3D shapes from partial 3D information
such as depth-maps. We report a lower mean CD of 0.089 on
test-instances from categories not included during training vs
0.0129 for PCN and 0.0124 for Folding-Net respectively.
\begin{figure}[!t]
\centering
\includegraphics[width=0.99\columnwidth]{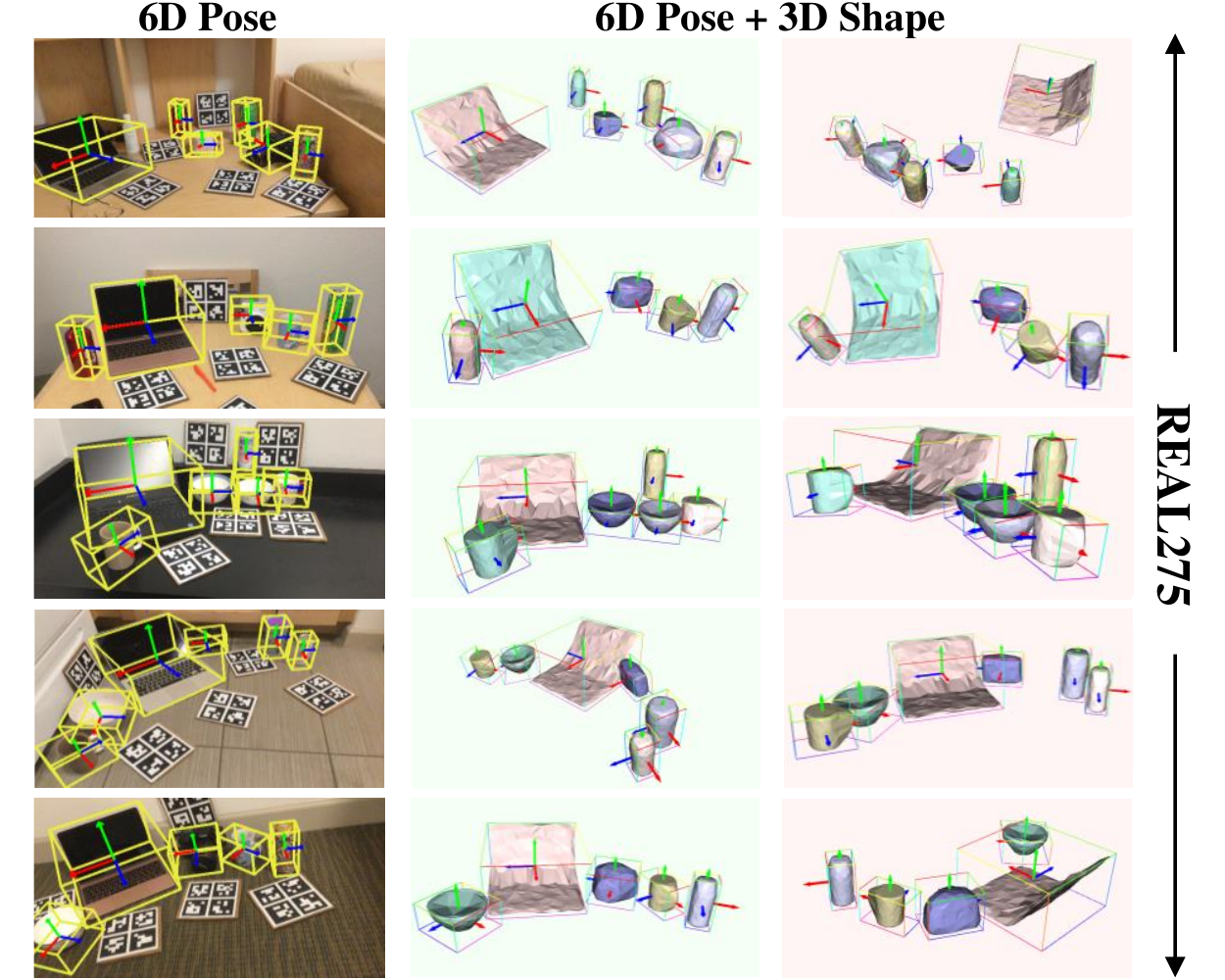}
\centering
  \caption{
  \textbf{Qualitative Results:} We visualize the real-world 3D shape prediction and 6D pose and size estimation of our method~(CenterSnap) from different viewpoints (green and red backgrounds).
  }
  \label{qualitative}
\end{figure} 
\textbf{Inference time:}
Given RGB-D images of size $640 \times 480$, our method performs fast~(real-time) joint reconstruction and pose and size estimation. We achieve an interactive rate of around 40 FPS for CenterSnap on a desktop with an Intel Xeon W-10855M@2.80GHz CPU and NVIDIA Quadro RTX 5000 GPU, which is fast enough for real-time applications. Specifically, our networks takes 22 ms and reconstruction takes around 3 ms. In comparison, on the same machine, competitive multi-stage baselines~\cite{wang2019normalized,tian2020shape} achieve an interactive rate of 4 FPS for pose estimation. \\
\renewcommand{\arraystretch}{1.0}
\begin{table}[t]
\scriptsize
\centering
\caption{\textbf{Ablation Study}: Study of the Proposed CenterSnap method on NOCS Real-test set to investigate the impact of different components i.e. Input, Shape, Training Regime~(TR) and Depth-Auxiliary loss~(D-Aux) on performance. C indicates Camera-train, R indicates Real-train and RF indicates Real-train with finetuning. 3D shape reconstruction evaluated with \textbf{CD} ($10^{-2}$). $*$ denotes the method does not evaluate size and scale and so has no IOU metric.}
% \zk{For ablation I would move (3) to second-to-last. Otherwise it's confusing what to get out of it (it seems C+RF is the dominant factor).
% Maybe also same for depth (2) as well. Otherwise it looks like going from D-> RGBD hurts performance}}
% \zk{Need arrows for higher/lower being better for each metric. You never mention this for lesser known ones such as CD, and it's not really clear (seems to be all over the place in ablation?)}}
% \zk{I think this is copied?}\zk{I assume below results aren't done yet? For example last three rows are the same results?}}
\vspace{+0.05cm}
\label{tab:ablation}
\resizebox{0.48\textwidth}{!}{%
\begin{tabular}{c@{\hskip 0.1in}lccccccccc} \\
\toprule
& & & & \multicolumn{6}{c}{\textbf{Metrics}} \\ 
\cmidrule(r{0.05in}){6-6}
\cmidrule(r{0.05in}){7-10}

& & & & & \textbf{3D Shape} & \multicolumn{4}{c}{\textbf{6D Pose}} \\ 
% \cmidrule(){5-9}
\cmidrule(r{0.05in}){6-6}
\cmidrule(r{0.05in}){7-10}
 {\textbf{\#}} & \text{Input} &\text{Shape} &\text{TR} & D-Aux&\textbf{CD}~$\downarrow$ & \textbf{IOU25}~$\uparrow$ & \textbf{IOU50}~$\uparrow$ &\textbf{5\textdegree \SI{10}{\cm}}~$\uparrow$ & \textbf{10\textdegree \SI{10}{\cm}}~$\uparrow$  \\

\midrule
 1 &RGB-D& \checkmark & C & \checkmark & 0.19 & 28.4 & 27.0 & 14.2 &48.2 \\
 2 &RGB-D& \checkmark & C+R & \checkmark & 0.19 & 41.5 & 40.1 & 27.1 & 58.2\\
 3 &RGB-D$^{*}$& &C+RF & \checkmark & --- & --- & --- & 13.8 & 50.2 \\
 4 & RGB & \checkmark & C+RF & \checkmark & 0.20 & 63.7 & 31.5 & 8.30 & 30.1\\
 5 &Depth& \checkmark & C+RF & \checkmark & 0.15 & 74.2 & 66.7 & 30.2 & 63.2  \\
 6 &RGB-D&\checkmark  & C+RF&  & 0.17 & 82.3 & 78.3 & 30.8 &68.3 \\
 7 &RGB-D& \checkmark & C+RF & \checkmark & 0.15 & 83.5 & 80.2 & 31.6 &70.9 \\
\bottomrule
\end{tabular}}
% \vspace{-1mm}
% \vspace{-0.15cm}
\end{table}
\textbf{Ablation Study:} An empirical study to validate the significance of different design choices and modalities in our proposed CenterSnap model was carried out. Our results are summarized in Table~\ref{tab:ablation}. We investigate the performance impact of \textbf{Input-modality}~(i.e. RGB, Depth or RGB-D), \textbf{Shape}, \textbf{Training-regime} and \textbf{Depth-Auxiliary loss} on the held out Real-275 set. Our ablations results show that our network with just mono-RGB sensor performs the worst~(31.5$\percent$ IOU50 and 30.1$\percent$ 6D pose at 10\textdegree \SI{10}{\cm}) likely because 2D-3D is an ill-posed problem and the task is 3D in nature. The networks with Depth-only~(66.7$\percent$ IOU50 and 63.2$\percent$ 6D pose at 10\textdegree \SI{10}{\cm}) and RGB-D~(80.2$\percent$ IOU50 and 70.9$\percent$ 6D pose at 10\textdegree \SI{10}{\cm}) perform much better. Our model without shape prediction under-performs the model with shape~(\#3 vs \#8 in Table~\ref{tab:ablation}), indicating shape understanding is needed to enable robust 6D pose estimation performance. The result without depth auxiliary loss~(0.17 CD, 78.3$\percent$ IOU50 and 68.3$\percent$ 6D pose at 10\textdegree \SI{10}{\cm}) indicates that adding a depth prediction task improved the performance of our model~(1.9$\percent$ absolute for IOU50 and 2.6$\percent$ absolute for 6D pose at 10\textdegree \SI{10}{\cm}) on real-world novel object instances. Our model trained on NOCS CAMERA-train with fine-tuning on Real-train~(80.2$\percent$ IOU50 and 70.9$\percent$ 6D pose at 10\textdegree \SI{10}{\cm}) outperforms all other training-regime ablations such as training only on CAMERA-train or combined CAMERA and REAL train-sets~(\#1 and- \#2 in Table~\ref{tab:ablation}) which indicates that sequential learning in this case leads to more robust sim2real transfer. \\
\textbf{Qualitative Results:} We qualitatively analyze the performance of CenterSnap on NOCS Real-275 test-set never seen during training. As shown in Figure~\ref{qualitative}, our method performs accurate 6D pose estimation and joint shape reconstruction on 5 different real-world scenes containing novel object instances. Our method also reconstructs complete 3D shapes~(visualized with two different camera viewpoints) with accurate aspect ratios and fine-grained geometric details such as mug-handle and can-head.
\section{Conclusion} Despite recent progress, existing categorical 6D pose and size estimation approaches suffer from high-computational cost and low performance. In this work, we propose an anchor-free and single-shot approach for holistic object-centric 3D scene-understanding from a single-view RGB-D. Our approach runs in real-time~(40 FPS) and performs accurate categorical pose and size estimation, achieving significant improvements against strong baselines on the NOCS REAL275 benchmark on novel object instances. 
\bibliography{bibliography.bib}
\end{document}